# Fast and Robust Structural Damage Analysis of Civil Infrastructure Using UAV Imagery

Alon Oring

Dynamic Infrastructure AI Research
alon@diglobal.tech

**Abstract.** The usage of Unmanned Aerial Vehicles (UAVs) in the context of structural health inspection is recently gaining tremendous popularity. Camera-mounted UAVs enable the fast acquisition of a large number of images often used for mapping, 3D model reconstruction, and as an assisting tool for inspectors. Due to the number of images captured during large scale UAV surveys, a manual image-based inspection analysis of entire assets cannot be efficiently performed by qualified engineers. Additionally, comparing defects to past inspections requires the retrieval of relevant images which is often impractical without extensive metadata or computer-vision-based algorithms.

In this paper, we propose an end-to-end method for automated structural inspection damage analysis. Using automated object detection and segmentation we accurately localize defects, bridge utilities and elements. Next, given the high overlap in UAV imagery, points of interest are extracted, and defects are located and matched throughout the image database, considerably reducing data redundancy while maintaining a detailed record of the defects.

Our technique not only enables fast and robust damage analysis of UAV imagery, as we show herein, but is also effective for analyzing manually acquired images.

**Keywords:** Predictive Maintenance, Preventive Maintenance, Deep Learning, Object Detection, Image Matching, Damage Assessment, Defect Matching.

# 1    Introduction

Transport infrastructure is crucial to the economic growth and social development of countries worldwide. Within this infrastructure, bridges are challenging structures to design, build and maintain due to a variety of loads and environmental conditions. For example: dead load, traffic, weather conditions, seismic events and long-term deterioration processes such as corrosion, wear and fatigue [1]. Prevention and control of degradation processes are achieved by implementing a maintenance plan taking into account the physical and environmental factors. Proper maintenance prevents damage and can increase the expected life of the bridge [22].

The European Construction Industry Federation (FIEC) warns about the threats of aging infrastructure as a large part of the critical infrastructure in EU MS (European Union Member States), especially bridges built in the post-war era with a design life of 50-100 years, and now facing ongoing deterioration. Studies have shown that if maintenance is neglected over a period of 3 years, it is estimated that the necessary repairs or renewals may cost 3 to 6 times more than the relevant timely maintenance [2].

Visual inspection is the primary method used to evaluate the condition of the majority of bridges worldwide [3,4]. It is common for initial inspections to find conditions that warrant repeat inspection and hence, repeated periodic visits are common. The process of physically tracking the progression of deficiencies is costly and time-consuming, especially when inspections must be carried out beneath the bridge deck, where special equipment would be needed to gain visual access for inspection.

In recent years, unmanned aerial vehicles (UAVs), also more commonly known as drones, have been used to monitor and inspect bridge infrastructure and have shown promise in efficient inspection of hazardous or hard to reach parts of bridges. Several studies have tested drone capabilities for bridge inspection, and it was concluded that they have some advantages over conventional inspection practices including cost, time, reduced risk for inspectors and inspection quality [5]. The current practice of visual inspection for bridges, as an initial diagnostic phase and as a recurring demand is difficult to perform in scale. Moreover, visual inspection results are often qualitative and subjective, leading to possible inconsistent reports [23].

Object detection algorithms offer a consistent and scalable approach to the analysis of visual inspection imagery. Object detection has attracted much research attention in recent years due to the significant advances in the design of deep learning network structures, the emergence of large-scale annotated training data and the developments in high-performance parallel computing systems. The performance of object detection algorithms in the inspection domain, however, has been relatively limited partly due to limited access to high-quality annotated data.

In this paper, we propose an end-to-end method for automated structural damage analysis of drone imagery. Our main contribution is a direct defect matching algorithm that enables the retrieval of defect occurrences throughout the image database, without relying on a 3D reconstruction of the asset.

## 2    Previous Works

The usage of drones in the context of structural health inspection is recently gaining tremendous popularity and many methods have been proposed in the past few years. The core requirement for a successful drone survey is the collection of multiple overlapping images of the bridge elements. This allows for the use of algorithms such as Structure from Motion (SfM) where the camera orientation and the geometry of the bridge elements can be used for the reconstruction of a 3D model which can be used as a permanent record of the geometry of the bridge. The model can then be used to allow navigation through the bridge and visual identification of defects. Using dense image matching, bundle adjustment and texture creation, fine-grained features of the bridge are reconstructed and made available for inspectors to analyze.

A common technique to view the components of the bridge is using an orthomosaic, which is a composite image made out of all images which have been orthorectified. The orthomosaic is a single image of the bridge element with no redundancy nor perspective. Performing defect and object recognition using the orthomosaic is a common approach [9,10,11,12]. For example, Ayele et al. [9] describe a data-driven modeling approach to UAV-assisted bridge inspections by performing automated crack segmentation on orthomosaic tiles. The benefit of this approach is straightforward as it reduces the redundancy of overlapping images for the object detection model. Performing defect detection on the original images can result in inconsistent predictions, where an instance of a defect is described by different bounding boxes or segmentation maps across multiple images.

The orthomosaic image generation can also introduce artifacts and distortions into the composite orthomosaic image, which in turn have a negative impact on the performance of object detection algorithms [6,7,8]. Additionally, when 3D reconstruction fails due to insufficient overlap, blurry images or partial cover of the element, defect detection cannot be efficiently performed due to high image redundancy and inference inconsistencies.

Seo et al. [13] proposed using a photogrammetry software to reconstruct a 3D model for damage observation. Manual inspection is performed on the 3D model and once a defect is selected, the photogrammetry software is used to retrieve the original images. While the retrieved unprocessed images allow for consistent detection, this method is not scalable due to the manual involvement during the initial 3D defect selection.

## 3   Proposed Method

We propose an end-to-end method for automated structural damage analysis where the original images are used for object detection, and direct defect matching is used to reduce data redundancy, without relying on 3D reconstruction or orthomosaic images.

### 3.1   Object Detection and Segmentation

The usage of object detection algorithms on the original drone imagery is conceptually simple and straightforward. This approach, however, creates an overabundance of predictions due to the overlap between images, thus leading to inconsistent predictions. Fig. 1 demonstrates predictions on three images. It can be seen that all three images capture the same region of the bridge, and object detection was performed for each image. Due to high data redundancy, the number of predictions will often be overwhelming for inspectors and asset managers to track and maintain. To manage all predictions efficiently, all occurrences of the same defect should be matched, as described in Section 3.2 and visualized by the colored line crossing the images in Fig. 1.

#### Model Training

The annotation methods and number of images used for each class are available in Table 1. For instance segmentation tasks we used Mask-RCNN (Region Based Convolutional Neural Networks) [14] and for semantic segmentation tasks we used DeepLab V3 [15]. Both networks were trained using the PyTorch framework [16] using 8 V100 GPUs (Graphics Processing Unit). During our experiments, we used an 80%-20% split for training and testing data, respectively. Images were gathered from inspection reports and drone imagery and were annotated in-house. After initial models were made available, erroneous and low certainty predictions were sent for re-labeling.

**Table 1.** Annotation details for all detection tasks.

| Category | Object | Annotation Method | Number of Images |
|----------|--------|-------------------|------------------|
| Defect | Crack | Semantic Segmentation | 10,000 |
| Defect | Corrosion | Semantic Segmentation | 10,000 |
| Elements | Column | Instance Segmentation | 2,000 |
| Utilities | Ruler | Instance Segmentation | 2,000 |

### 3.2   Defect Matching

Following the above challenges in handling multiple defects in overlapping images, we propose a new method called *defect matching*, which finds all occurrences of the same defect throughout the image database. Our approach comprises an image retrieval step that finds overlapping images and a defect matching step that matches between different occurrences of the same defect in overlapping images.

**Content-Based Image Retrieval**

Content based image retrieval has been an outstanding research topic in the computer vision society. The two dominant image retrieval methods are SIFT-based and CNN-based [21]. The Scale-Invariant Feature Transform (SIFT) based methods mostly rely on the Bag of Words (BoW) model which uses the SIFT descriptors to compute a single vector per image, which can then be used for retrieval based on some similarity metric. In recent years, the popularity of CNN (Convolutional Neural Networks) based methods increased due to the hierarchical structure that has been shown to outperform hand-crafted features in many vision tasks. The CNN-based retrieval models usually compute vector representations using a neural network and calculate the similarity between image pairs.

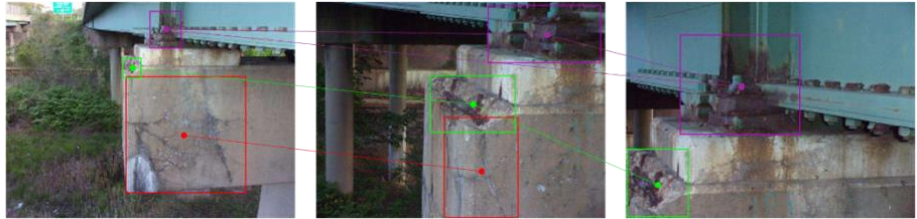

**Fig. 1.** Defect matching example. Different defect types are drawn in different colors. Defect matching is performed between all relevant image pairs. Multiple instances of the same defect were correctly matched across different drone images. Green bounding box: exposed reinforcement defect correctly detected across multiple images. Defect matching is represented by the lines between different occurrences of a specific defect.

Using each method separately for image retrieval in drone imagery may result in false retrievals, as demonstrated in Fig. 2. Due to the similarity between the rulers, feature matching techniques might yield a false positive prediction of similarity between the images, since the region captured is different. Combining both methods, together with filtering utilities such as rulers, allows us to reduce the number of false matches. We used both similarity scores as follows:

$$S(I_1,I_2) = \alpha \cdot S_{SIFT}(I_1,I_2) + (1-\alpha) \cdot S_{CNN}(I_1,I_2) \qquad (1)$$

Where $S_{SIFT}, S_{CNN}$ are the similarity scores between the images $I_1,I_2$ for the SIFT and CNN vectors, respectively, and $0 \leq \alpha \leq 1$.

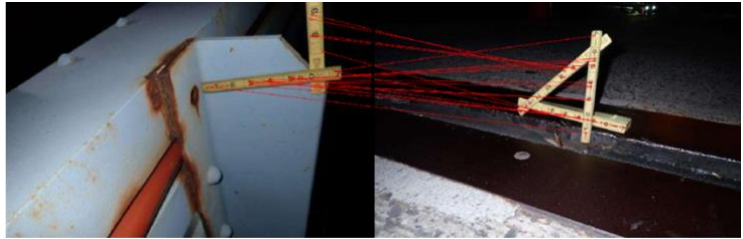

**Fig. 2.** False image retrieval due to a ruler. Red lines indicate feature matching caused by the ruler marks and numbers.

**Defect Matching Algorithm**

Once correct matches between images have been found, different occurrences of the same defect should be matched. Defect matching is performed by extracting a large number of features using techniques such as Scale-Invariant Feature Transform (SIFT) [17], Speeded Up Robust Features (SURF) [18] and Oriented FAST and rotated BRIEF (ORB) [19] from each image. The features are then matched using a feature matching algorithm such as Brute-Force matching or Fast Library for Approximate Nearest Neighbors (FLANN) [20]. Feature extraction and feature matching can be calculated directly without a full 3D reconstruction of the asset and without prior requirements common with 3D reconstruction software such as a minimal number of images.

The result of the feature matching stage is a list of matched keypoints between two images where every keypoint is characterized by the 2D position in the image and the features associated with the keypoint local neighborhood. Every pair of matched keypoints represent the same object in two different images. Next, we iterate over all matched keypoints between a pair of images. A pair of matched keypoints between two images defines a single 2D coordinate per image. If both 2D coordinates reside within the boundaries of a defect prediction of the same class, we mark this keypoint match as valid. Finally, if the number of valid keypoint matches between two defect predictions crosses a threshold, those defects are matched. This process is repeated for all matched features across all pairs of retrieved images.

As can be seen in Fig. 1, once defect matching has been established across the image dataset, every defect is associated with all its occurrences. For example, the purple bounding box in Fig. 1 corresponds to corrosion. Without defect matching, three separate occurrences of the same instance of corrosion will be reported. However, using the aforementioned method, all occurrences of the defect are matched across all images capturing it and only a single defect, spanning across multiple images, will be reported.

## 4 Results

### 4.1 Object Detection & Segmentation

Evaluation metrics for the instance and semantic segmentation tasks are available in Tables 2,3, respectively. Inference examples are available in Fig. 3. All metrics and visualizations are reported on a testing dataset. As seen in both qualitative and quantitative analysis, we reach high values of accuracy and average precision across multiple defects, elements and utilities.

**Table 2.** Elements and utilities instance segmentation evaluation results.

| Object | AP | AP50 | AP75 | APs | APm | APl |
|--------|--------|--------|--------|--------|--------|--------|
| Column | 56.421 | 77.377 | 60.710 | 37.669 | 50.510 | 73.642 |
| Ruler | 55.514 | 86.132 | 62.072 | 22.222 | 53.343 | 59.647 |

**Table 3.** Defect semantic segmentation evaluation results.

| Object | IoU | ACC |
|----------|--------|--------|
| Crack | 60.403 | 70.567 |
| Corrosion | 54.116 | 55.248 |

Additionally, we compare our models to the performance of inspection experts, and evaluate whether or not we are able to correctly detect the defects described in inspection reports, which are prepared and authorized by qualified civil engineers. Our methodology consists of using the detailed description available in the inspection reports and comparing it to the inference results manually. We used 3 experts that were presented with the image, the report description and the model prediction. Each expert determined if the model prediction included the defect detailed in the inspection report. The final decision was determined using a majority vote between the experts. The recall of our models, when compared to inspection experts across all defect classes, is 90.64%.

### 4.2 Defect Matching

Evaluation of defect matching is challenging since all occurrences of the same defects throughout the entire image dataset must be determined manually, which is time-consuming and labor-intensive. During ground truth curation, we manually performed defect matching on 3 datasets of different substructures of a bridge, captured during a drone survey. Each dataset contains 150 images.

We performed evaluation as follows: we obtained all pairwise defect matches in the ground truth and compared them to the retrieved pairwise defect matches. Evaluation metrics for the pairwise analysis are available in Table 4. While this method correctly evaluates our algorithm performance generally, it fails to capture the civil engineering point-of-view, which is matching all occurrences of a defect, regardless if all pairwise matches are found. We propose an evaluation approach based on a *chain* of matched defects. A defect chain is defined using a connected graph where the nodes are the defects and an edge between defects exists if there is a match between the defects. The defect chain is the set of nodes of such connected graph. Evaluating chains allows us to measure the number of defects in each chain, rather than the number of pairwise matches.

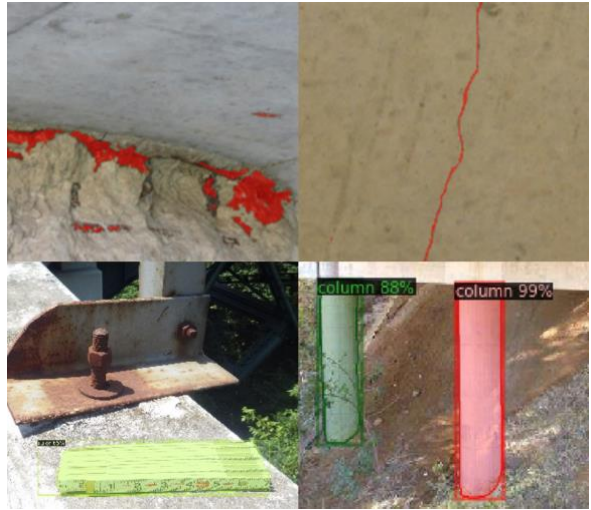

**Fig. 3.** Top row: semantic segmentation of corrosion (left) and cracks (right). Bottom row: instance segmentation of rulers (left) and columns (right).

For example, Fig. 1 demonstrates defect matching of the green defect between the leftmost image and the middle image, and between the middle image and the rightmost image. In this example, the defect matching between the leftmost and the rightmost images failed while the purple defect is matched across all possible image combinations. In this example, the pairwise metric score for the green defect matching will be lower than the chain metric score.

Chain evaluation is performed as follows: we calculate the intersection between the predicted chain and all ground-truth chains. If the intersection between the chains contains at least two defects and is at least half the size of the ground-truth chain, we count that as a true positive. If no such chain is found, the predicted chain counts as false positive. Once a ground-truth chain is assigned to a predicted chain, it is removed from the list of possible ground-truth candidates. We also test our matching approach on non-UAV imagery using a dataset of 82 images that were taken from inspection reports performed over multiple years. Our results are presented in Table 4.

The decrease in the precision metric in the drone chain when compared to pairwise could be explained as follows: If a predicted chain crosses the intersection threshold, the corresponding ground-truth chain is removed from the ground-truth candidates. Thus, the chain metric penalizes instances where a single chain is split into multiple chains.

**Table 4.** Defect matching results.

| Image Type | Metric | Precision | Recall |
|------------|--------|-----------|--------|
| UAV | Pairwise | 100.00 | 86.25 |
| UAV | Chain | 54.28 | 81.42 |
| non-UAV | Pairwise | 42.10 | 19.20 |
| non-UAV | Chain | 48.87 | 32.78 |

The results for the non-UAV imagery suggest we retrieved nearly 33% of the defect chains automatically. The low recall for the pairwise matches is likely due to the difference between images taken by human inspectors. Such images are often taken from various perspectives, lighting conditions and in different times, thus posing a challenge on both image retrieval algorithms and feature matching. The chain metrics are better than the pairwise metrics in the non-UAV imagery due to positive chains being found using few pairwise matches, and fewer instances of correct chains being split into multiple chains. Automatic non-UAV defect matching is demonstrated in Fig. 4. Note the changes in lighting conditions that is not common in drone imagery.

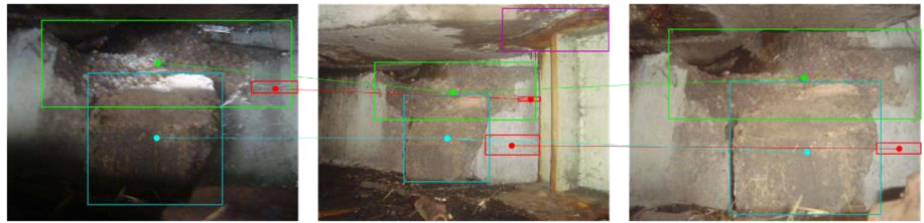

**Fig. 4.** Defect matching example. Different defect types are drawn in different colors. Defect matching is performed between all relevant image pairs. Multiple instances of the same defect were correctly matched across different non-UAV images. Red bounding boxes represent detected cracks. Red lines between images indicate that the defects were matched.

## 5    Conclusion & Discussion

The problem of robust defect analysis in inspection imagery has been discussed in depth in the past few years. Nevertheless, it is our opinion that reconstructing a 3D model of the asset as a preprocessing step for object detection and defect matching has a negative effect on the usage of drones in bridge inspection. In addition, the image requirements associated with 3D model reconstruction does not support defect tracking and matching of non-UAV acquired images.

The performance of object detection algorithms is highly affected by the quality of the training dataset. Curating a civil engineering defect database is a costly and complex process. Even when several qualified engineers are performing image annotation, a consensus regarding the annotation might not be agreed upon. Additionally, inspectors from different countries with different regulations often annotate defects differently. We believe curating a large, high-quality database will drive the research endeavors of future detection and matching algorithms.

In this paper, we proposed an end-to-end method for automated structural damage analysis of drone imagery. We showed that our process enables fast and robust analysis, without reconstructing a 3D model of the asset, which in turn allows simpler drone data acquisition. Additionally, we demonstrated that our defect matching technique is also effective when analyzing manually acquired images and can be used for the analysis of inspection imagery from different years, multiple sensors, perspectives and lighting conditions.